\documentclass[11pt,a4paper]{article}
\usepackage[hyperref]{emnlp2020}
\usepackage{amsmath,amsfonts,bm}

\def\eqref#1{equation~\ref{#1}}
\def\1{\bm{1}}

\def\va{{\bm{a}}}
\def\vb{{\bm{b}}}

\def\vd{{\bm{d}}}

\def\vs{{\bm{s}}}

\def\vu{{\bm{u}}}

\def\vx{{\bm{x}}}

\def\va{{\bm{a}}}
\def\vb{{\bm{b}}}

\def\vd{{\bm{d}}}

\def\vs{{\bm{s}}}

\def\vu{{\bm{u}}}

\def\vx{{\bm{x}}}

\def\vpi{{\bm{\pi}}}

\def\mV{{\bm{V}}}

\DeclareMathAlphabet{\mathsfit}{\encodingdefault}{\sfdefault}{m}{sl}
\SetMathAlphabet{\mathsfit}{bold}{\encodingdefault}{\sfdefault}{bx}{n}

\def\gL{{\mathcal{L}}}

\newcommand{\softmax}{\mathrm{softmax}}

\DeclareMathOperator*{\argmax}{arg\,max}

\usepackage{times}
\usepackage{latexsym}
\usepackage{hyperref}
\usepackage{url}
\usepackage{float}
\usepackage{booktabs}
\usepackage{multirow}
\usepackage{graphicx}
\usepackage{boxedminipage}
\usepackage{xcolor}
\usepackage{subcaption}
\usepackage{colortbl}
\usepackage{cleveref}

\definecolor{gblue}{RGB}{66,133,244}
\Crefname{equation}{Eq.}{Eqns.}
\Crefname{figure}{Fig.}{Figs.}

\usepackage{microtype}

\aclfinalcopy % Uncomment this line for the final submission
\title{TED: A Pretrained Unsupervised Summarization Model \\ with Theme Modeling and Denoising}

\author{Ziyi Yang$^1$\thanks{$\;\;$Equal contribution. Work was done during first author's internship at Microsoft.}$\;$, Chenguang Zhu$^{2*}$, Robert Gmyr$^2$, Michael Zeng$^2$, Xuedong Huang$^2$, Eric Darve$^1$ \\
Stanford University$^1$\\
\texttt{\{zy99,darve\}@stanford.edu}\\
Microsoft Cognitive Services Research Group$^2$\\
\texttt{\{chezhu,rogmyr,nzeng,xdh\}@microsoft.com} \\
}

\date{}

\begin{document}
\maketitle
\begin{abstract}
Text summarization aims to extract essential information from a piece of text and transform the text into a concise version. Existing unsupervised abstractive summarization models leverage recurrent neural networks framework while the recently proposed transformer exhibits much more capability. Moreover, most of previous summarization models ignore abundant unlabeled corpora resources available for pretraining. In order to address these issues, we propose TED, a transformer-based \textbf{unsupervised abstractive} summarization system with pretraining on large-scale data. We first leverage the lead bias in news articles to pretrain the model on millions of unlabeled corpora. Next, we finetune TED on target domains through theme modeling and a denoising autoencoder to enhance the quality of generated summaries. Notably, TED outperforms all unsupervised abstractive baselines on NYT, CNN/DM and English Gigaword datasets with various document styles. Further analysis shows that the summaries generated by TED are highly abstractive, and each component in the objective function of TED is highly effective.
\end{abstract}

\section{Introduction}
Summarization refers to the task of condensing a document into a shorter version without losing the key information. Summarization models can be categorized into two types: abstractive and extractive. Extractive models select sentences from the input article as the summary. Such process ensures a basic level of grammaticality and accuracy, but also limits the model ability to copying. In contrast, abstractive models summarize a document using newly generated tokens and phrases that may not be found in the original article, which involves a process requiring an advanced ability to refine, paraphrase and re-organize language information \citep{see2017get, narayan2018ranking,knowsum}.

Like most machine learning algorithms, summarization models can also be divided into supervised and unsupervised categories. Supervised approaches require in-domain parallel data, i.e. both input articles and corresponding reference summaries must be present for the teacher-forcing training \citep{hermann2015teaching, bertsum}. Unfortunately, high-quality paired data are not always available across different text domains and styles. Moreover, considering the fact that summarization is not an easy task even for people, reliable human-labeled data are also difficult to obtain. Therefore, several unsupervised summarization approaches have been proposed, which do not require reference summaries for the target domain. We introduce these methods as follows.

\textbf{Unsupervised extractive models.} TextRank \citep{mihalcea2004textrank} encodes sentences in the article as nodes in an undirected graph. The weights of edges are measured by sentences similarity. The centrality of a node (sentence) is computed by PageRank \citep{pagerank} to decide whether a sentence should be included in the final summary. \citet{zheng2019sentence} advances upon TextRank by encoding sentences with BERT representation \citep{devlin2018bert} to compute pairs similarity and build graphs with directed edges decided by the relative positions of sentences.

\textbf{Unsupervised abstractive models.} \citet{baziotis2019seq} leverages differentiable sampling and optimizes by re-constructing the input article from the generated summary. \citet{meansum} proposes a similar idea in the multi-document summarization setting. \citet{wang2018learning} uses adversarial training and reinforcement learning to make the summary human-readable. \citet{fevry} adopts denoising autoencoders originally used in sentence compression. However, most of these models are only tested on datasets with considerably small article/summary length. Also, previous models usually utilize the recurrent neural networks (RNNs). However, transformers \citep{NIPS2017_7181, devlin2018bert} have shown superior performances over RNNs on various NLP tasks, including machine translation, reading comprehension, sentiment analysis, etc. Few Efforts have been made to leverage transformers in unsupervised abstractive summarizations.

\textbf{Pretraining Language Model.} 
In recent years, pretraining language models have proved to be quite powerful in solving numerous NLP tasks. The state-of-the-art pretrained models include CoVe \citep{cove}, ELMo \citep{elmo}, GPT \citep{gpt}, BERT \citep{devlin2018bert} and UniLM \citep{unilm}. Taking advantage of corpora with billions of tokens, the pretrained language models learn universal and robust representations for various semantic structures and linguistic relationships. As a result, pretrained models have been widely used with considerable success in applications such as question answering \citep{sdnet}, sentiment analysis \citep{elmo} and passage reranking \citep{reranking}. Furthermore, UniLM \citep{unilm} leverages its sequence-to-sequence capability for abstractive summarization; the BERT model has been employed as an encoder in BERTSUM \citep{bertsum} for supervised extractive and abstractive summarization. 

In this paper, we present TED, a pretrained \textbf{unsupervised abstractive} summarization model which is finetuned with theme modeling and denoising on in-domain data. TED utilizes a transformer-based encoder-decoder structure and the pretraining leverages large-scale corpora containing millions of unlabeled articles. Our primary contributions are two-fold as follows.

First, we leverage the lead bias in news articles to pretrain TED. The lead bias is introduced by the journalistic convention of writing using an inverted pyramid structure, placing the most important information in the beginning of an article. We propose to use the leading sentences as the target summary and train the model to predict it during pretraining. In this way, we pretrain a summarization model on a large-scale corpus with 21.4M news articles. The model yields better performance than most existing unsupervised methods.

Second, to finetune on specific datasets, TED is further trained with a theme modeling loss and a denoising autoencoder. The role of the theme modeling module is to make the generated summary semantically close to the article. The module uses a semantic classifier trained using a discriminative objective function. Furthermore, to optimize on the generated summary tokens, we adopt the Gumbel-Softmax \citep{jang2016categorical} estimator to replace the non-differentiable $\argmax$.
The denoising autoencoder has been previously used in unsupervised machine translation \citep{lample2017unsupervised} and sentence compression \citep{fevry}, and we employ it to help the model extract salient information from corrupted text.

Instead of classical word tokenization, we adopt the SentencePiece tokenization \citep{sp} to alleviates the long-standing out-of-vocabulary (OOV) problem in language generation tasks \citep{oov, oov1}. We test TED on several benchmark datasets. The experimental results show that TED outperforms all unsupervised abstractive baselines on all datasets. For example, on the CNN/DM dataset, it outperforms the state-of-the-art unsupervised abstractive model by more than 9 ROUGE-1 points and compares favorably with most unsupervised extractive models. We further show that TED is capable of generating novel words and phrases in summaries and is a highly abstractive system even compared with supervised systems.

\section{Methodology}
\begin{figure*}[htbp]
\centering
\includegraphics[width=0.75\textwidth]{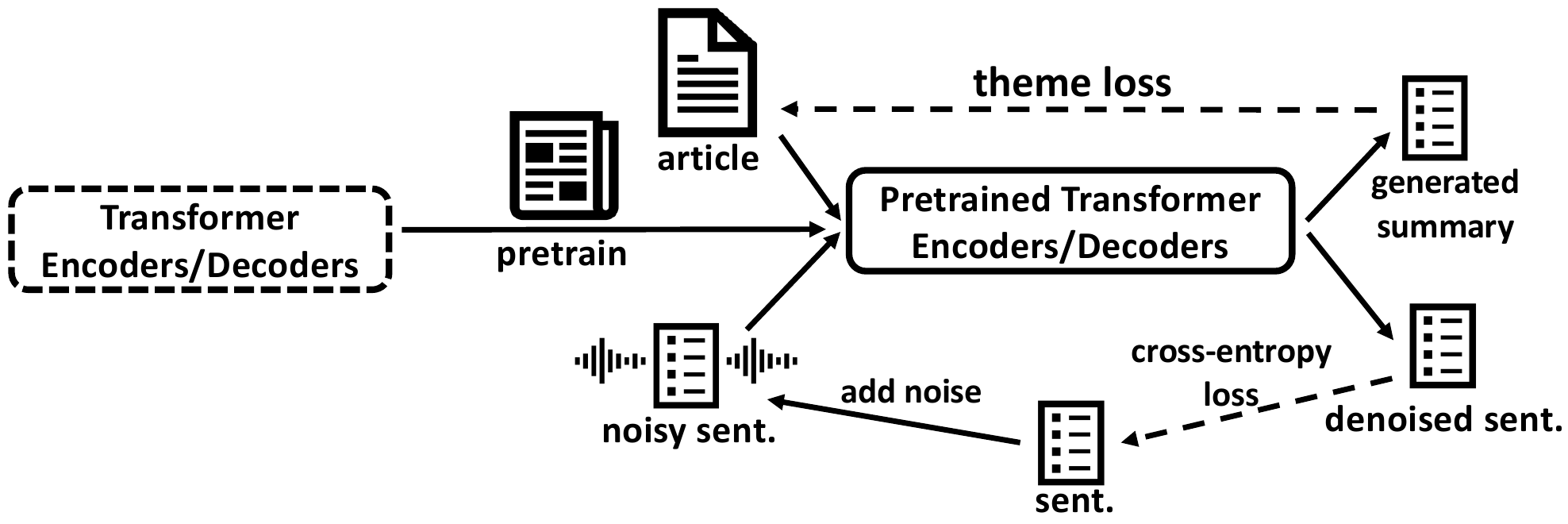}
\caption{Overall structure of our model. TED first pretrains on news articles and then finetunes with theme modeling and denoising. (from left to right).}
\label{fig:model}
\end{figure*}
In this section, we will go through the model structure of TED, i.e. the transformer encoder and decoder. Then we introduce the pretraining method and two in-domain finetuning objectives: theme modelling and the denoising autoencoder. The overall architecture of TED is illustrated in \Cref{fig:model}.

\subsection{Transformer Encoder and Decoder}

Previous unsupervised summarization methods are based on the sequence to sequence (seq2seq) model \citep{seq2seq} that primarily uses the RNN model. As the transformer structure \citep{NIPS2017_7181} has been successfully applied in a large number of NLP tasks, TED employs the multi-layer transformer encoder-decoder architecture. We follow the standard transformer design in TED networks and refer readers to \citet{NIPS2017_7181} for more technical details on transformers. Denote the number of layers (i.e., Transformer blocks) as $L$, the number of self-attention heads as $H$ and the hidden size as $N$. We explore two different configurations in experiments, 4 layers 4 heads (4L4H) with $N = 512$ and 10 layers 8 heads (10L8H) with $N = 720$.

Denote the input article tokens sequence as $X = \{x_1, x_2, ..., x_n\}$, and each token is first transferred to a vector by a trainable embeddings matrix $\mV$. The output from transformer encoder $E$ is a sequence of encoded vectors $E(X) = \{\vu_1^E, \vu_2^E, ..., \vu_n^E\}$. The decoder can be viewed as a conditional language model to generate the summary depending on the generator outputs. Given $k$ input summary tokens $W = \{w_1, w_2, ..., w_k\}$, the cross attention layer in the decoder $D$ attends with encoder outputs $\{\vu^E_i\}_{i=1}^{n}$. The decoder outputs are $D(\{w_1, w_2, ..., w_k\}) = \{\vu_1^D, \vu_2^D, ..., \vu_k^D\}$. The probability distribution over the vocabulary for $w_{k+1}$ is given by:
\begin{equation}
P(w_{k+1}|w_{1:k}, x_{1:n})=\softmax(\mV \vu^D_{k})
\label{eq:decoder}
\end{equation}

In traditional tokenization algorithms, efforts have been made to address the out-of-vocabulary (OOV) issue \citep{kg2vec} at the cost of losing semantic information, such as mapping OOV words to a special ``UNK'' token. To mitigate the open vocabulary problem, we adopt SentencePiece \citep{sp}, a data-driven method that trains tokenization models from sentences in large-scale corpora. The advantage of the SentencePiece model is that its subwords can cover all possible word forms and the subword vocabulary size is controllable. In the evaluation experiments, we train a SentencePiece subword vocabulary of size 32,000.

Note for supervised summarization models, during training, the inputs to the decoder are the groundtruths/reference summary tokens; for unsupervised learning, input tokens are generated in the previous pass, i.e. one new token is generated in one pass. More details are available in \cref{sec:gumbel}.

\subsection{Pretraining with Unlabeled Corpora}
Leveraging large scale unlabeled text corpora to pretrain models has been proven as an effective method in multiple NLP tasks \citep{devlin2018bert}. However, such approach has not yet been utilized in text summarization.

News articles follow an inverted pyramid structure, i.e. front loading the most salient information. This so-called "lead bias" for news summarization is so strong that \citet{see2017get} have shown that using the first 3 sentences in a news article as a summary can score higher than many sophisticated deep learning models. Although this poses a great challenge to previous research, we take advantage of this property in our favor in the pretraining phase of TED.

For a news article, we set the target summary to be the first three sentences. This allows the model to exploit the structural bias of the news domain and infer the most important information using the background materials in the remainder of the article. To collect data for pretraining, we obtain three years of online news articles from 2016 to 2019 via an industrial search engine. The search engine indexes major online news domain, for instance, New York Times and Bloomberg. Then we collect the parsed articles within the 2016-2019 time range as the raw data. Note that this time span does not overlap any of three test datasets we use in this paper, therefore the pretraining should not lead to data leakage in test. It is also worth noting that this idea of utilizing structural bias for large-scale summarization pretraining is not limited to specific types of models, and it can be applied to other types of text as well: academic papers with abstracts, novels with editor's notes, books with tables of contents.

\begin{figure}[tbp]
\centering
  \includegraphics[width=0.755\columnwidth]{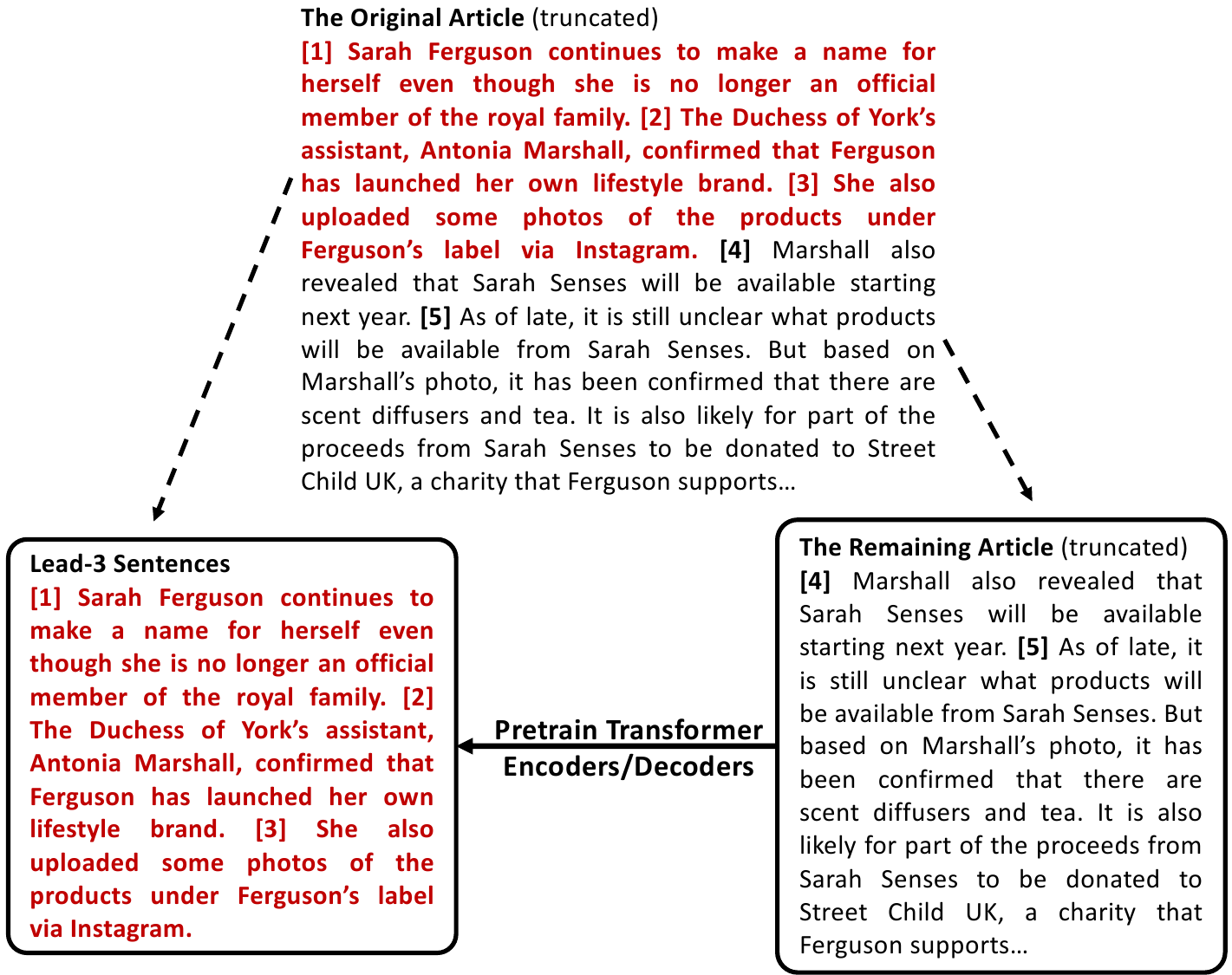}
  \caption{An example of the pretraining task: predict the Lead-3 sentences (as the target summary) using the rest of the article.}
  \label{fig:top3}
\end{figure}

However, one should carefully examine and clean the source data to take advantage of lead bias, as the top three sentences may not always form a good summary. Therefore, we conduct strict data cleaning to remove irrelevant distracting content and filter out articles whose top three sentences do not form a good summary:

First, many news articles begin with media names, reporter names, dates or other irrelevant information for summarization, e.g. ``New York (CNN) --'', ``Adam Smith, June 3rd 2018:''. We automatically clean these using regular expressions.

Second, we only include articles whose top three sentences contain between 10 and 150 words, and remaining sentences contain between 150 and 1,200 words. The criterion on top three sentences is set to filter out articles with either extremely short leading sentences, e.g. phrases of one or two words, which contain too little information to be reasonable summaries, or exceedingly long leading sentences to reduce the pretraining time. The limit on total number of words in the article is to filter out very long articles to reduce memory consumption. Another purpose is to remove very short articles of which the information is too condensed and not suitable for summarization pretraining.

Third, we also remove articles in which the first three sentences may not contain the major information in the article. We use a simple and easy-to-compute metric: overlapping words. We compute the portion of non-stopping words in the top three sentences that also appear in the rest of an article. A higher ratio indicates that the rest of the article is likely to elaborate on the beginning part. We keep those articles with the ratio of overlapping words higher than 0.65. We pick this threshold based on observations in the CNN/DM dataset, where the median overlapping ratio of non-stopping words between golden summary and the article is 0.87, and the median ratio between the top three sentences and the rest of the article is 0.77. Setting the threshold at 0.65 makes the final training set size fit with the available computation resources and ensures that the leading sentences contain enough information.

Finally, we end up with 21.4M articles, out of which 12,000 articles are randomly sampled as the validation set. We conduct pretraining for 10 epochs and pick the model with the best ROUGE-L score on the validation set. The pretraining task is to predict to the first three sentences of an article using the \textbf{rest of the article} (so pretraining will not teach the model to simply copy the leading three sentences since they are removed from the input to the transformers). Note that TED does not start off from other pretrained models like Bert.

After pretraining, in order to adapt TED to a specific target dataset (for evaluation), we finetune TED on the target dataset in an unsupervised manner. The finetuning objective functions includes the following: \textit{theme modeling} and \textit{denoising autoencoder}.

\subsection{Theme Modeling}
Theme modeling aims to make the generated summary semantically close to the input article. We employ differential sampling to enable optimization on generated summaries and train a classifier to improve the semantic relatedness between the output summary and article.

\subsubsection{Differentiable Sampling}
\label{sec:gumbel}
In order to optimize the transformers using output summaries, we need to make the generation of summary tokens differentiable. Recall the conditional probability distribution of token $w_{k+1}$ is $P(w_{k+1}|w_{1:k}, x_{1:n})=\softmax(\mV \vu^D_{k})$. Let $\vpi$ denote $P(w_{k+1}|w_{1:k}, x_{1:n})$. One can use $\argmax$ on $\vpi$ to obtain the token $w_{k+1}$ in the forward pass, however, it is not differentiable in the gradient back-propagation. Although $\argmax$ can be avoided by obtaining the embedding of $w_{k+1}$ as a weighted sum of the vocabulary embeddings $\mV$, this results in an undesirable gap between the training (weighted sum) and the inference (discrete sampling) on the forward pass generation. To solve this issue, we employ the straight-through Gumbel-Softmax estimator \citep{jang2016categorical} as in \citet{style, baziotis2019seq}. Specifically, the forward pass in training still uses $\argmax$ sampling, but for gradient computation, the following Gumbel-Softmax distribution is used as a differentiable approximation for the $\argmax$ operation:

\begin{equation}
    \bm{\tilde{\pi}_i} = \frac{\exp(\log(\vpi_i) + g_i)/\tau)}{\sum_{j = 1}^k\exp(\log(\vpi_j) + g_j)/\tau)}
\end{equation}
where $g_1, \cdots, g_k$ are i.i.d samples drawn from the Gumbel distribution $G(0, 1)$ and $\tau$ denotes the softmax temperature. As shown in \citet{jang2016categorical}, as $\tau \to 0$, the Gumbel-Softmax distribution converges to the categorical (one-hot) distribution; as $\tau \to \inf$, the Gumbel-Softmax distribution converges to the uniform distribution. Although this gradient estimator is biased, we find that this method works well in practice. We choose $\tau = 0.1$ based on the CNN/DM validation set and use this value in all the experiments. Denote the input article as $\vd$, the generated summary as $\vs = \{w_1, w_2, ..., w_m\}$. The generation of $\vs$ follows the recursive process that input $w_{1:k}$ to the transformer decoder to obtain $w_{k+1}$, then input $w_{1:k+1}$ to compute $w_{k+2}$ and so on. The first input token $w_{1}$ is always the special beginning token \texttt{[START]}.

\subsubsection{Encoder Transformer as A Semantic Classifier}
As the generated summary may be off the article theme at the beginning of finetuning, we also optimize TED such that the generated summaries are semantically closed to the input articles. We frame the semantic similarity problem in a discriminative setting. To better adapt to the target-domain data, we add sentence pairs from training articles to facilitate similarity computation.

\begin{figure}
\centering
\includegraphics[width=1\columnwidth]{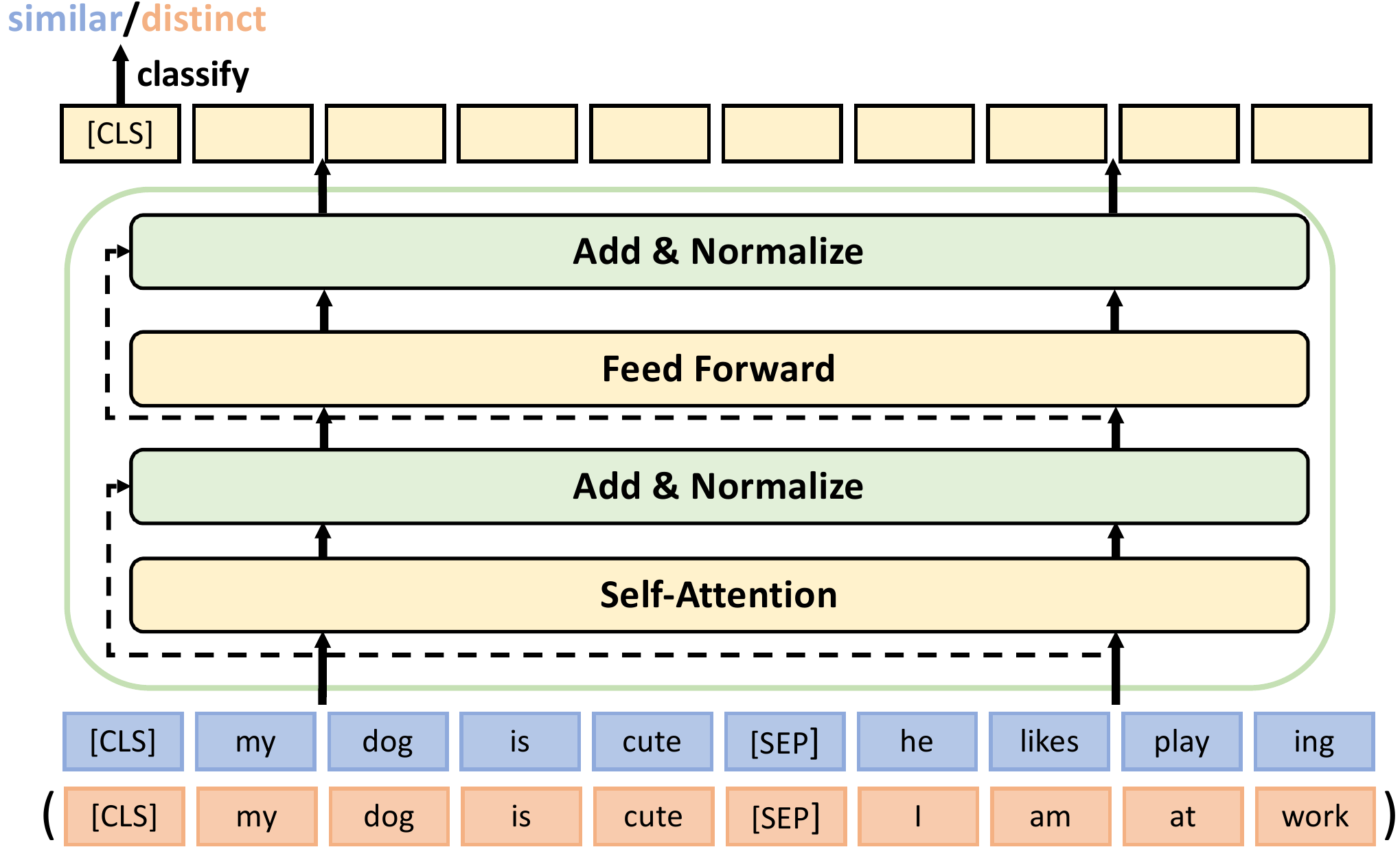}
\caption{Theme modeling is essentially updating TED with a semantic classifier. The input sentence pair is first processed by adding a ``class'' token in the beginning and a ``separation'' token between the two sentences. Then the sentence pair is fed into the transformer encoder, and the first output vector is classified to ``similar'' or ``distinct''.}
\label{fig:thm}
\end{figure}

Concretely, during training, we pick two consecutive sequences of tokens $\va_1$ and $\va_2$ from an article to form a positive sequence pair \{$\va_1$, $\va_2$\}. Second, sequence $\vb_1$ is chosen from another random article in the dataset to form the negative sequence pair \{$\va_1$, $\vb_1$\}. Following \citet{devlin2018bert}, each sequence pair is packed into one single sequence by inserting a special token \texttt{[SEP]} between them and adding trainable segment embeddings. A special classification token \texttt{[CLS]} is also added to the beginning of the packed sequence. As shown in \Cref{fig:thm}, the packed sequence is then fed as input into TED's transformer encoder. The output vector associated with the token \texttt{[CLS]}, is then classified into similar/distinct categories by a two-layer fully connected network. We use the following cross-entropy loss to optimize the encoder such that the $\va_1$ is semantically similar to $\va_2$ and $\vs$ is also closed to $\vd$, while $\va_1$ is semantically distinct from $\vb_1$.

\begin{equation}
\begin{split}
\gL_{theme} &= -\log(p(y = 1|\va_1, \va_2))\\
-&\log(p(y = 1|\vs, \vd)) -\log(p(y = 0|\va_1, \vb_1))
\label{eq:theme}
\end{split}
\end{equation}

\subsection{Denoising Autoencoder}
The idea of denoising autoencoder \citep{dae} has been used in unsupervised machine translation \citep{artetxe2017unsupervised, lample2017unsupervised} to prevent the model from learning to merely copy every input word one by one. This denoising process imitates text simplification and helps to refine essential semantic information.

In detail, a sequence of $n$ consecutive tokens $\vx$ from the input article is injected with two types of noise. First, we insert noisy tokens sampled from other articles in the same dataset into the original sequence at random positions, obtaining a new sequence with length $n'$, where $n'$ is $40\%$-$50\%$ larger than $n$. Next, similar to \citet{lample2017unsupervised}, the sequence is slightly shuffled by applying a permutation $\sigma$ such that $\forall i \in [1,2,\dotsi,n'], |\sigma(i) - i| \leq k$, where the permutation distance $k$ is set to be $20\%$ of the length of $\vx$. The final corrupted sequence is denoted as $\vx'$. TED model is then trained to recover the original token sequence given the corrupted sequence:

\begin{equation}
\gL_{denoise} = CE(\vx, \mbox{TED}(\vx'))
\label{eq:denoise}
\end{equation}

where $CE$ denotes the mean of token-level cross-entropy loss. $\mbox{TED}(\vx')$ denotes the sequence of probability distribution outputs $\{\vpi\}$ from the decoder with inputting $\vx'$ to the encoder. The final objective function is the mean of \Cref{eq:theme} and \Cref{eq:denoise} (we empirically find that equal weights between the two terms work well enough in practice):
\begin{equation}
    \gL_{\mbox{TED}}=\frac{\gL_{theme}+\gL_{denoise}}{2}
\end{equation}

It is worth pointing out that we do not conduct ``pretraining'' on target evaluation datasets. This is because for a target dataset, we do not know beforehand whether the Lead-X sentences will make a quality summary or not. We do have the option to do so on datasets where Lead-X are good summaries, however, it is potentially cherry-picking datasets. Also, we do not conduct supervised finetuning with ground-truths summaries in evaluation datasets because we want to have an entirely unsupervised summarization system with motivations stated in the introduction section.

\section{Experiments}
\subsection{Datasets}
We evaluate our model on three benchmark summarization datasets: NYT, CNN/DM and English Gigaword, containing 110K, 300K and 3.8M news articles, respectively. The detailed statistic information on the datasets can be found in the appendix. In NYT, following \citet{bertsum}, we choose 4,000 examples as the validation set and filter out examples with summaries of fewer than 50 words. In CNN/DM, similar to \citet{see2017get} and \citet{bertsum}, input articles are truncated to 500 tokens. In English Gigaword, we filter out data examples with articles containing only”UNK” tokens.

%. The text statistics on these datasets are summarized in \Cref{tab:data}. Numbers of NYT and CNN/DM are collected from \citet{zheng2019sentence}.

\subsection{Baseline and Metrics}
We compare TED with the following baselines. (1) Unsupervised abstractive systems: Brief \citep{wang2018learning}, SEQ$^{3}$ \citep{baziotis2019seq}, GPT-2 (\citet{gpt2}, without supervised finetuning with ground-truths summaries). (2) Unsupervised extractive systems: TextRank \citep{mihalcea2004textrank}, Lead-X. (3) Supervised abstractive and abstractive (models trained with ground-truths summaries): PACSUM \citep{zheng2019sentence}, PGNet \citep{see2017get}, REFRESH \citep{narayan2018ranking} and SUMO \citep{sumo}. TED is \textbf{unsupervised abstractive} and therefore not directly comparable with supervised baselines. The purpose of supervised systems here is for 
references. We describe the implementation details of our model in Appendix. We measure the quality of generated summaries by ROUGE F1 score \citep{lin2004rouge}, including unigram (ROUGE-1), bigram (ROUGE-2) and longest common subsequence (ROUGE-L).

\subsection{Results}
Results on English Gigaword dataset are shown in \Cref{tab:exp}, TED outperforms all unsupervised baselines. \Cref{tab:exp} shows the experimental results on NYT and CNN/DM datasets. In NYT, the unsupervised fine-tuning of TED improves upon the pretrained model by 2.75\%/1.06\%/2.37\% on ROUGE-1/ROUGE-2/ROUGE-L respectively. Note that ROUGE metric prefers extractive systems that preserve original phrasing \citep{see2017get}. Considering this factor, TED achieves results that are competitive with unsupervised extractive baselines and surpasses all unsupervised abstractive models. In CNN/DM, TED with a larger model size (10L8H) outperforms all unsupervised abstractive methods and compares favorably with unsupervised extractive baselines. Note that TED outperforms GPT-2, a powerful transformer-based language generation model pretrained on large scale webpage textual data, by significant margins. Again, TED further improves upon pretrained models on both 10L8H and 4L4H configurations.

\begin{table}[htbp]
\caption{Results on the English Gigaword dataset. Performances of baseline models are collected from their original papers. The best performance in each metric is in bold.}
\begin{center}
\scalebox{0.9}{
\begin{tabular}{l|ccc}
 \toprule
 Model & R1 & R2 & RL \\
 \midrule
 TED 10L8H (ours) & \textbf{25.58} & \textbf{8.94} & \textbf{22.83} \\

 Pretrained 10L8H (ours) & 25.23 & 8.84 & 22.56 \\

 TED 4L4H (ours) & 24.59 & 8.10 & 21.91 \\
 Pretrained 4L4H (ours) & 22.52 & 7.46 & 20.09 \\
 LEAD-8 & 21.86 & 7.66 & 20.45\\
 SEQ$^{3}$ & 25.39 & 8.21 & 22.68 \\
 Brief & 21.26 & 5.60 & 18.89 \\
\bottomrule
\end{tabular}
}
\label{table:giga}
\end{center}
\end{table}

\begin{table*}[htbp]
\caption{ROUGE $F_1$ scores on CNN/DM and NYT datasets. R1/R2/RL stands for ROUGE-1/ROUGE-2/ROUGE-L respectively. Best results in each unsupervised category is in bold. Results of baseline models are obtained from their original papers or running open-sourced codes.}
\begin{center}
\begin{tabular}{l|ccc|ccc}
 \toprule
  & \multicolumn{3}{c}{CNN/DM} & \multicolumn{3}{c}{NYT}\\
  \midrule
  Model & R1 & R2 & RL & R1 & R2 & RL\\
  \hline
  \rowcolor[gray]{0.95}\multicolumn{7}{c}{\textit{Unsupervised Abstractive}} \\
  \hline
  TED 10L8H (ours) & \textbf{38.73} & \textbf{16.84} & \textbf{35.40} & \textbf{37.78} & \textbf{17.63} & \textbf{34.33} \\
  Pretrained 10L8H (ours) & 38.38 & 16.49 & 35.08 & 35.03 & 16.57 & 31.96 \\
  TED 4L4H (ours) & 34.38 & 9.56 & 30.10 & 24.45 & 7.97 & 21.77  \\
  Pretrained 4L4H (ours) & 31.20 & 10.05 & 27.80 & 22.56 & 7.38 & 18.79\\
  SEQ$^3$ & 23.24 & 7.10 & 22.15 & 17.85 & 3.94 & 19.53\\
  Brief & 28.11 & 9.97 & 25.41 & - & - & - \\
  GPT-2 & 29.34 & 8.27 & 26.58 & - & - & -  \\
  \hline
  \rowcolor[gray]{0.95}\multicolumn{7}{c}{\textit{Unsupervised Extractive}} \\
  \hline
  LEAD-3 & 40.50 & 17.70 & 36.70 & 35.50 & 17.20 & 32.00\\
  TextRank + tf-idf & 33.20 & 11.80 & 29.60 & 33.20 & 13.10 & 29.00\\
  TextRank + skip-thought & 31.40 & 10.20 & 28.20 & 30.10 & 9.60 & 26.10 \\
  TextRank + BERT & 30.80 & 9.60 & 27.40 & 29.70 & 9.00 & 25.30\\
  PACSUM + tf-idf & 39.20 & 16.30 & 35.30 & 40.40 & 20.60 & 36.40 \\
  PACSUM + skip-thought & 38.60 & 16.10 & 34.90 & 38.30 & 18.80 & 34.50  \\
  PACSUM + BERT & \textbf{40.70} & \textbf{17.80} & \textbf{36.90} & \textbf{41.40} & \textbf{21.70} & \textbf{37.50} \\
 \hline
 \rowcolor[gray]{0.95}\multicolumn{7}{c}{\textit{Supervised Abstractive \& Extractive}} \\
 \hline
 SUMO  & 41.00 & \textbf{18.40} & 37.20 & 42.30 & \textbf{22.70} & \textbf{38.60} \\
 PGNet & 39.50 & 17.30 & 36.40 & \textbf{42.70} & 22.10 & 38.00 \\
 REFRESH & \textbf{41.30} & \textbf{18.40} & \textbf{37.50} & 41.30 & 22.00 & 37.80 \\
\bottomrule
\end{tabular}
\label{tab:exp}
\end{center}
\end{table*}

\begin{figure*}[htbp]
\centering
 \begin{boxedminipage}{1\textwidth}
 %\small
 \textbf{Article:}  \newline
 after exposing potential security risks with airlines' in-flight entertainment systems, one of the top experts on counter-threat intelligence in the world was pulled off a flight by fbi agents. chris roberts, who featured in a string of fox news reports, was yanked off his plane after it landed in syracuse, new york, on wednesday night by two fbi agents and two uniformed officers. roberts, who works for security intelligence company one world labs, was questioned for the next four hours ...\newline
 \textbf{TED Summary:} \newline
 chris roberts, who works for security intelligence company one world labs, was pulled off a plane in syracuse, new york, on wednesday night by two fbi agents and two uniformed officers. the incident occurred only a few hours after a report about roberts' research was released by the government accountability office earlier this week.\newline
 \textbf{Reference:}  \newline
 chris roberts of one world labs grabbed after plane landed in syracuse. two fbi agents spent four hours questioning him about cyberhacking. agents confiscated electronic devices and computer files from roberts.
 he flew in to give talk at aerospace conference about plane vulnerabilities. roberts featured on fox news' on the record with greta van susteren. regarded as one of the world's top experts on counter-threat intelligence."
 \end{boxedminipage}
 \caption{An example of a generated summary by TED. The reference summary and parts of the input article are also included.}
 \label{fig_abs_ref_exs}
\end{figure*}

\section{Discussion}
\subsection{Ablation Study}

The ablation studies shown in \Cref{table:abl} verify the effectiveness of each component in TED. Training the transformer encoder-decoder from scratch yields reasonable performance. Pretraining on large-scale data results in more than 10\% improvement on all three metrics on training TED from scratch. Pretraining plus either theme modeling or denoising improves upon the pretrained model by more than 2\%. The full TED model, pretraining with theme modeling and denoising, produces the best result overall.
\begin{table}[htbp]
\begin{center}

\caption{Ablation study of different components in TED on the NYT dataset. We test with the 10L8H model configuration.}
\scalebox{0.85}{
\begin{tabular}{l|ccc}
 \toprule
 Model & R1 & R2 & RL \\
 \midrule
 Train from scratch & 24.49 & 4.41 & 20.14\\
 Pretrained only & 35.03 & 16.57 & 31.96 \\
 Pretrained w/ theme modeling & 37.16 & 18.18 & 34.15 \\
 Pretrained w/ denoise loss & 37.48 & 17.83 & 34.05 \\
 Full model & 37.78 & 17.63 & 34.33 \\
\bottomrule
\end{tabular}}
\label{table:abl}
\end{center}
\end{table}

\begin{figure}[htbp]
\centering
\includegraphics[width=0.4\textwidth]{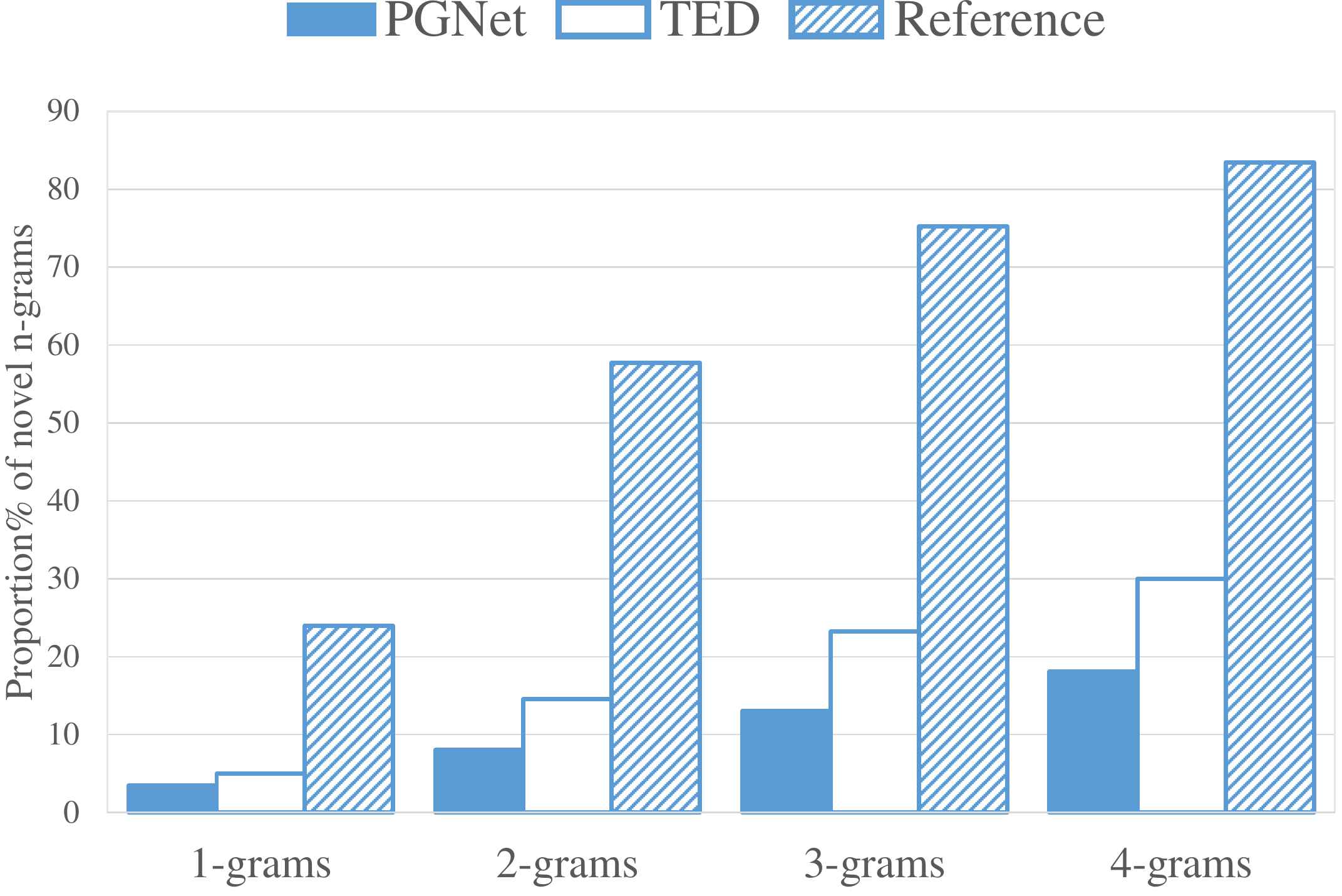}
\caption{Proportion of novel grams in summaries generated by different models on the CNN/DM test set.}
\label{fig:novel}
\end{figure}

\subsection{Model Analysis}

\textbf{Example.} We showcase a sample summary from CNN/DM dataset along with the input article and the reference summary (\Cref{fig_abs_ref_exs}). As shown, TED is able to capture and organize the essential information into fluent and highly readable language. We attribute the grammatical correctness to the pretraining process and the denoising autoencoder. However, we also note that although TED manages to recognize the temporal information related to reported event (a few hours after Fox news reports), it makes a mistake by summarizing as ``a few hours after a report about roberts' research was released\dots''. It shows that fact cross-checking is a potential future research direction.

\textbf{Abstractiveness.} To examine how abstractive TED is, we compute the proportion of novel N-grams in the summary output (\Cref{fig:novel}). The reference summary and the output from PGNet are included for comparison. Although TED is unsupervised, it includes more novel grams than the supervised model PGNet. The reference summaries have the highest proportion of n-grams.

\subsection{Comparison with Previous Unsupervised Models}
TED is an innovative unsupervised summarization model with several distinctive features setting it apart from previous approaches such as MeanSum and SEQ$^3$. First, TED leverages the structure of news articles for an effective large-scale pretraining. Second, although both MeanSum and SEQ$^3$ have a loss to make the summary similar to the input article, they leverage the classical cosine similarity on text embeddings. In contrast, TED innovatively encodes the similarity by a transformer encoder with much more modeling capability. Third, the denoising module in TED is completely distinct from the idea of reconstruction in SEQ$^3$ and MeanSum. In TED’s denoising module, the corrupted texts are input to the transformer and the model is trained to filter the added noises. The original clean document is not used as input and thus unseen by TED in the forward pass. However, the reconstruction process in MeanSum and SEQ$^3$ employs the original document to generate a summary, which is then used to reconstruct the original document.

\section{Conclusion}
In this paper, we propose TED, an unsupervised abstractive summarization model. First, we introduce an effective large-scale pretraining approach leveraging the lead bias in news articles. The pretraining employs automatic filtering mechanism and does require any human-labeled data. We then develop a finetuning scheme to induce the semantic similarity between summaries and input articles, together with a denoising autoencoder to improve the quality of generated summaries. Experiments across three datasets show that TED significantly outperforms unsupervised abstractive baselines.

\bibliographystyle{acl_natbib}
\bibliography{anthology,emnlp2020}

\clearpage
\appendix
\section{Implementation Details}
For pretraining, we use a dropout rate of 0.3 for all inputs to transformer layers. We use RAdam \citep{radam} as the optimizer, with a learning rate of $10^{-4}$. Also, due to the different numerical scales of the positional embedding and initialized sentence piece embeddings, we divide the positional embedding by 100 before feeding it into the transformer. We pretrain one model for 10 epochs. After each epoch, the model is evaluated on validation data. We pick the check points with the highest ROUGE L.

For unsupervised finetuning on specific datasets, the learning rate is set to $2\times10^{-4}$ and dropout ratio stays the same as in pretraining. The batch size is 16, and the vocabulary embeddings are also updated in the training process. During the test phase, we generate the summarization from trained encoder and decoder by beam search. The ROUGE version we use for evaluation is ROUGE-1.5.5. This is consistent with benchmark models whose version of ROUGE are available in open-sourced codes and original papers.

At test time, we limit the longest length of generated summaries, which is set based on validation dataset. For instance, the maximum generation length for CNN/DM dataset is 175.

\section{Datasets Information}
For a better understanding of the evaluation protocols, the statistical information of evaluation datasets is summarized in \Cref{tab:data}.
\begin{table}[h]
\centering
\caption{Average document and summary length in number of words and sentences on NYT, CNN/DM, and English Gigaword datasets (test set).}
\scalebox{0.85}{
\begin{tabular}{l|c|cc|cc}
\toprule
\multirow{2}{*}{{Dataset}}  & \multirow{2}{*}{{ \# docs}} &  \multicolumn{2}{c|}{{ avg. document}} & \multicolumn{2}{c}{{ avg. summ.}} \\
&  &  words   &  sen. & words & sen.  \\
\midrule

CNN/DM           &  11,490 & 641.9   & 28.0 & 54.6   & 3.9  \\
NYT              &  4,375  & 1,290.5 & 50.7 & 79.8   & 3.5 \\
Gigaword &  1,937  & 29 & 1 & 8 & 1 \\
\bottomrule
\end{tabular}
}
\label{tab:data}
\end{table}
\end{document}